\documentclass{article}

\usepackage[final]{NFFL2021}

\usepackage[utf8]{inputenc} 
\usepackage[T1]{fontenc}    
\usepackage{hyperref}       
\usepackage{url}            
\usepackage{booktabs}       
\usepackage{amsfonts}       
\usepackage{nicefrac}       
\usepackage{microtype}      
\usepackage{xcolor}         

\makeatletter
\newcommand{\verbatimfont}[1]{\renewcommand{\verbatim@font}{\ttfamily#1}}
\makeatother
\makeatletter
 \let\Ginclude@graphics\@org@Ginclude@graphics 
\makeatother
\usepackage{amsmath}
\usepackage{tcolorbox}

\usepackage{algorithm}
\usepackage{algpseudocode}
\usepackage{varwidth}
\usepackage{fancyvrb}
\usepackage{graphicx}
\usepackage{paralist}

\newcommand{\fedjax}{\textsc{FedJAX}}

\newcommand{\neurips}[1]{#1}
\newcommand{\arxiv}[1]{}

\title{\textsc{FedJAX}: Federated learning simulation with JAX}

\author{%
  Jae Hun Ro \\
  Google Research \\
  \texttt{jaero@google.com} \\
  \And
  Ananda Theertha Suresh \\
  Google Research \\
  \texttt{theertha@google.com} \\
  \And 
  Ke Wu \\
  Google Research \\
  \texttt{wuke@google.com} \\
}

\begin{document}

\maketitle

\begin{abstract}
Federated learning is a machine learning technique that enables training across decentralized data. Recently, federated learning has become an active area of research due to an increased focus on privacy and security. In light of this, a variety of open source federated learning libraries have been developed and released. 
We introduce \fedjax{}, a JAX-based open source library for federated learning simulations that emphasizes ease-of-use in research. With its simple primitives for implementing federated learning algorithms, prepackaged datasets, models and algorithms, and fast simulation speed, \fedjax{} aims to make developing and evaluating federated algorithms faster and easier for researchers. Our benchmark results show that \fedjax\ can be used to train models with federated averaging on the EMNIST dataset in a few minutes and the Stack Overflow dataset in roughly an hour with standard hyperparameters using TPUs.
\end{abstract}

\section{Introduction}

\begin{figure}[t]
\centering
\includegraphics[scale=0.4]{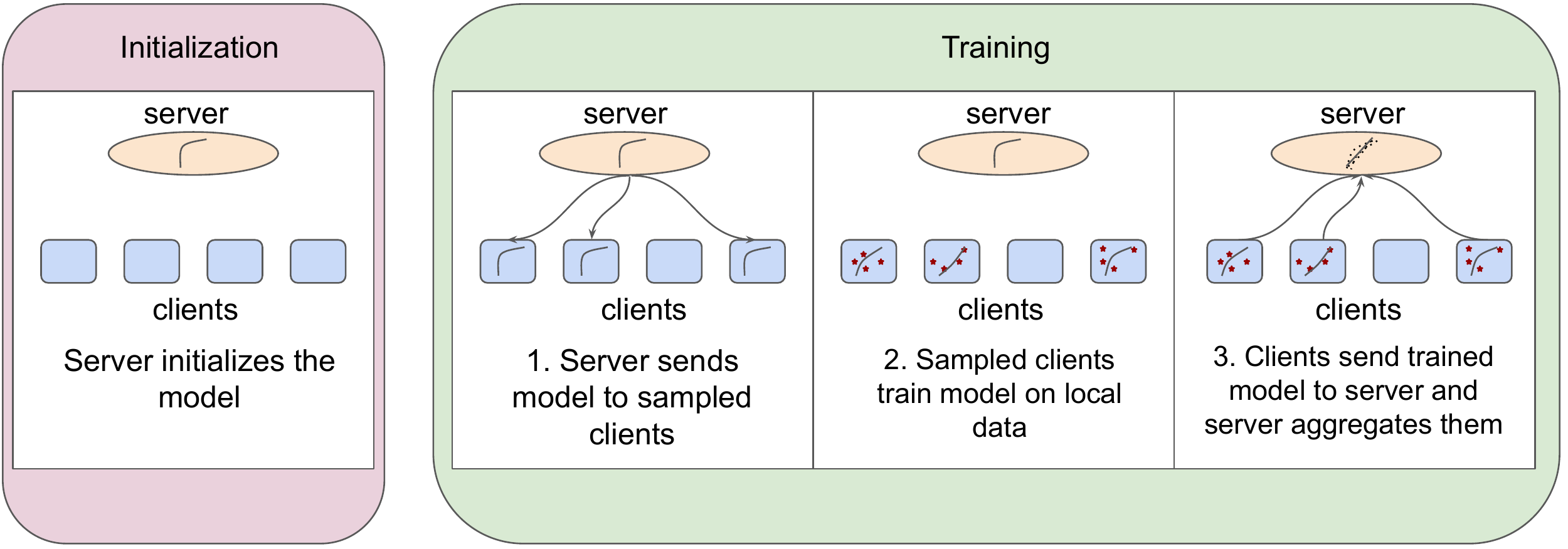}
\caption{An example federated learning algorithm with four clients.}
\label{fig:fedlearning}
\end{figure}

Federated learning is a machine learning setting where many clients collaboratively train a model under the orchestration of a central server, while keeping the training data decentralized. Clients can be either mobile devices or whole organizations depending on the task at hand \citep{konevcny2016federated,konecny2016federated2,mcmahan2017communication, yang2019federated}. Federated learning is typically studied in two scenarios: \emph{cross-silo} and \emph{cross-device}. In cross-silo federated learning, the number of clients is small, where as in cross-device, the number of clients is very large and can be in the order of millions. Figure~\ref{fig:fedlearning} highlights key  characteristics of most federated learning algorithms in the cross-device settings.
Typically, federated learning algorithms first initialize the model at the server and then complete three key steps for each round of training: 
\begin{enumerate}
\item The server selects a subset of clients to participate in training and sends the model to these clients.
\item Each selected client completes some steps of training on their local data. 
\item After training, the clients send their updated models to the server and the server aggregates them together.
\end{enumerate}
For example, Algorithm~\ref{alg:FedAvg} illustrates the popular \emph{federated averaging} algorithm \citep{mcmahan2017communication}, which follows the above three steps. Federated learning has demonstrated usefulness in a variety of contexts, including next word prediction \citep{hard2018federated, yang2018applied} and healthcare applications \citep{brisimi2018federated}. We refer to \citep{li2019federatedsurvey, kairouz2021advances} for a more detailed survey of federated learning.

Federated learning poses several interesting challenges. For example, training typically occurs mostly on small devices, limiting the size of models that can be trained.  Furthermore, the devices may have low communication bandwidth and require model updates to be compressed. Data is also distributed in a non i.i.d. fashion across devices which raises several optimization questions. Finally, privacy and security are of utmost importance in federated learning and addressing them requires techniques ranging from differential privacy to cryptography.

Given these challenges, federated learning has become an increasingly active area of research. This includes new learning scenarios \citep{MohriSivekSuresh2019,abay2020mitigating}, optimization algorithms \citep{li2018fedprox,yu2019parallel,li2019fedconvergence,haddadpour2019convergence,khaled2020tighter, karimireddy2019scaffold, ro2021communication}, compression algorithms \citep{suresh2017distributed, caldas2018expanding, xu2020ternary}, differentially private algorithms \citep{agarwal2018cpsgd, peterson2019private, sattler2019clustered}, cryptography techniques \citep{bonawitz2017practical}, and algorithms that incorporate fairness \citep{Li2020Fair, du2020fairnessaware, huang2020fairness}. Motivated by this, there are several libraries for federated learning, including, TensorFlow Federated \citep{tff}, PySyft \citep{ryffel2018generic}, FedML \citep{chaoyanghe2020fedml}, FedTorch \citep{ibmfl2020ibm}, and Flower \citep{ibmfl2020ibm}.

Recently, JAX \citep{jax2018github} was introduced to provide utilities to convert Python functions into Accelerated Linear Algebra (XLA) optimized kernels, where compilation and automatic differentiation can be composed arbitrarily. This enables expressiveness for sophisticated algorithms and efficient performance without leaving Python. Given its ease-of-use, several libraries to support machine learning have been built on top of JAX, including, but not limited to, Flax \citep{flax2020github}, Objax \citep{objax2020github}, Jraph \citep{jraph2020github}, and Haiku \citep{haiku2020github} for neural network architectures and training, Optax \citep{optax2020github} for optimizers, Chex \citep{chex2020github} for testing, and RLax \citep{rlax2020github} for reinforcement learning.

We present \fedjax, a JAX and Python based library for federated learning simulation for research. 
\fedjax\ is designed for ease-of-use for research and is not intended to be deployed over distributed devices. Focusing on ease-of-use, the \fedjax\ API is structured to reduce the amount of new concepts that users have to learn to get started and comes packaged with several standard datasets, models, and algorithms that can be used straight out of the box. Additionally, since it is based on JAX, \fedjax\ can run on accelerators (GPU and TPU) with minimal additional effort.

The rest of the paper is organized as follows. In Section~\ref{sec:design}, we overview the system design, in Section~\ref{sec:examples}, we demonstrate a sample federated learning algorithm with \fedjax, and in Section~\ref{sec:experiments}, we benchmark training with \fedjax{} on two datasets. 

\section{System design}
\label{sec:design}

A typical federated learning experiment consists of a federated dataset, model and optimizer, local client training strategy, and server aggregation strategy, so we structure \fedjax{} accordingly. 
The design was primarily driven by ease-of-use and performance when addressing the challenges uniquely attributed to using JAX for federated learning.
\subsection{Datasets and models}

\paragraph{Federated dataset}


In the context of federated learning, data is decentralized and distributed across clients, with each client having their own local set of examples. We refer to two levels of organization for datasets:

\begin{itemize}
    \item Federated dataset: A collection of clients, each with their own local dataset and metadata.
    \item Client dataset: The set of local examples for a particular client.
\end{itemize}

\begin{figure}
    \centering
    \includegraphics[scale=0.42]{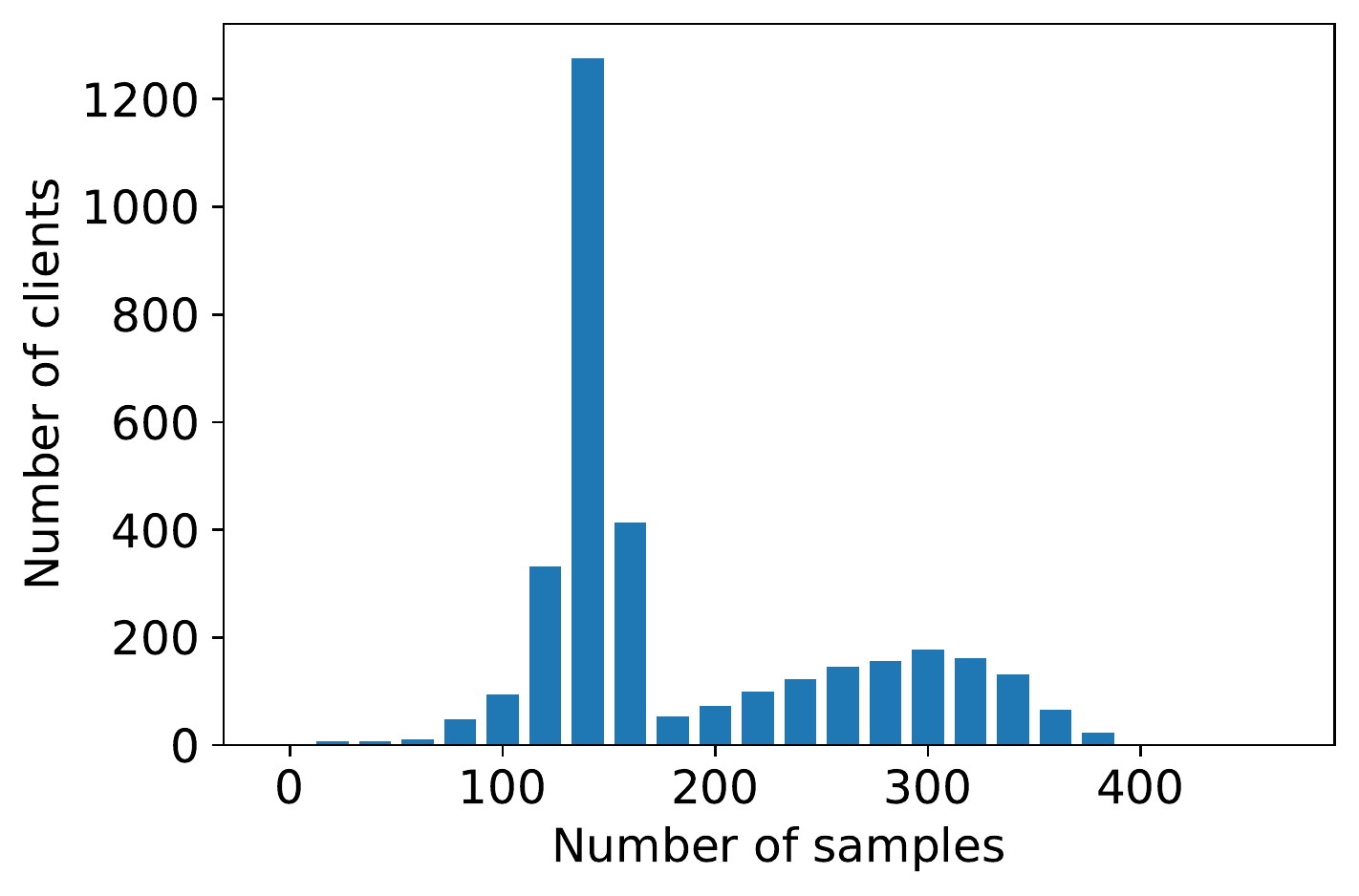} \hspace{0.1\linewidth}
    \includegraphics[scale=0.42]{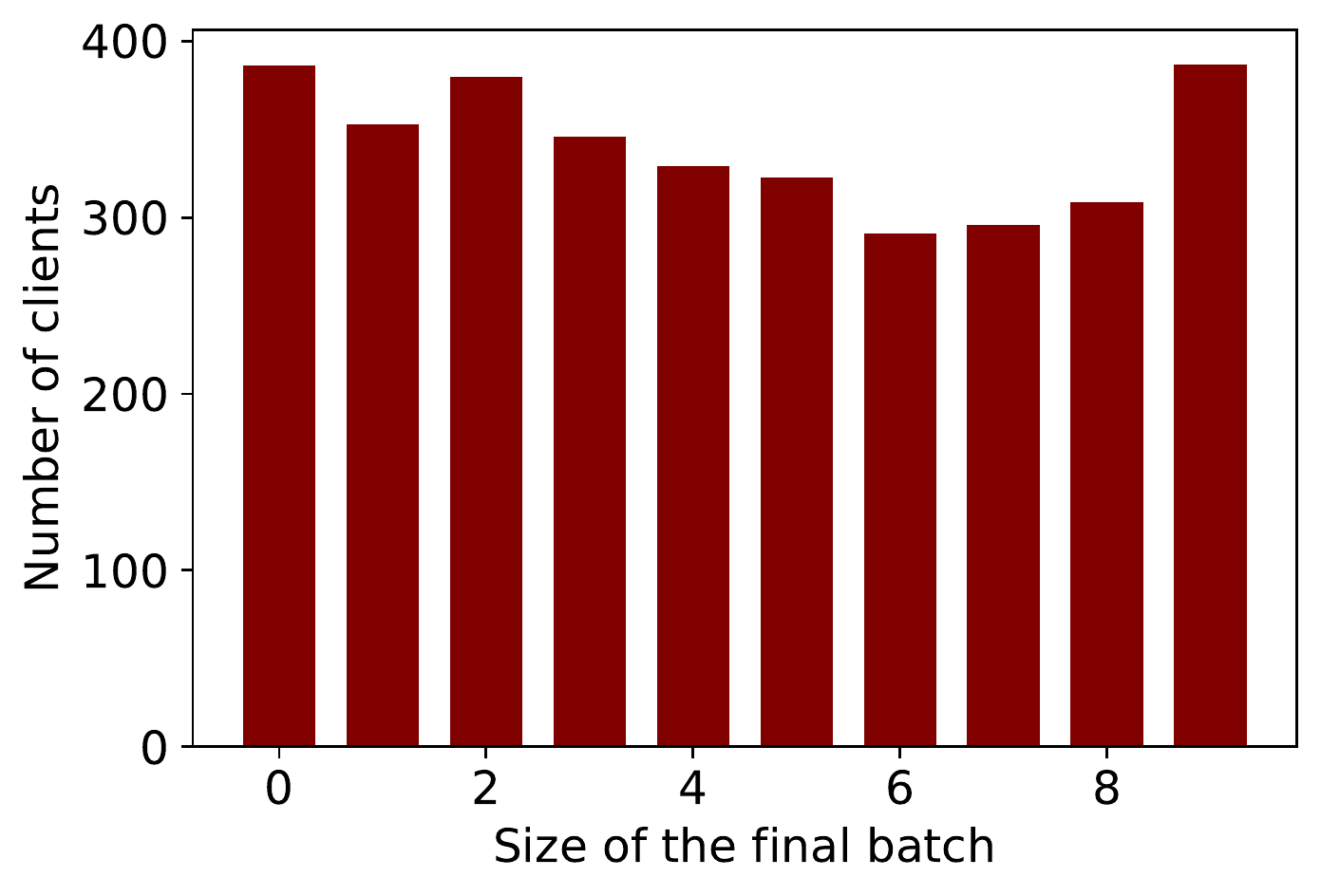}
    \caption{EMNIST dataset characteristics. Left: Histogram of number of clients as a function of number of samples. Right: Histogram of number of clients as a function of the size of the final batch, with batch size $10$.}
    \label{fig:emnist-hist}
\end{figure}

In its simplest form, federated datasets are just mappings from clients to their local examples.
Specifically, clients have a unique identifier for querying their local dataset, which is essentially treated as a list of examples. Furthermore, these local client datasets are typically small as seen in the EMNIST example in Figure~\ref{fig:emnist-hist}. This is in contrast to standard centralized machine learning which requires iterating over a single  dataset in a large number of batches. With this difference in mind, we designed \texttt{fedjax.FederatedData} and \texttt{fedjax.ClientDataset}
to rely mostly on NumPy and Python, making it
 easy to use and troubleshoot. Finally, in order to take full  advantage of JAX, we also provide several helpful functions for accessing and iterating over federated and client datasets. \neurips{We refer readers to the tutorial\footnote{\url{https://fedjax.readthedocs.io/en/latest/notebooks/dataset_tutorial.html}} for an overview of these functionalities.}

%
%
%

\begin{figure}
    \centering
    \begin{tabular}{c c c}
      \includegraphics[scale=0.35]{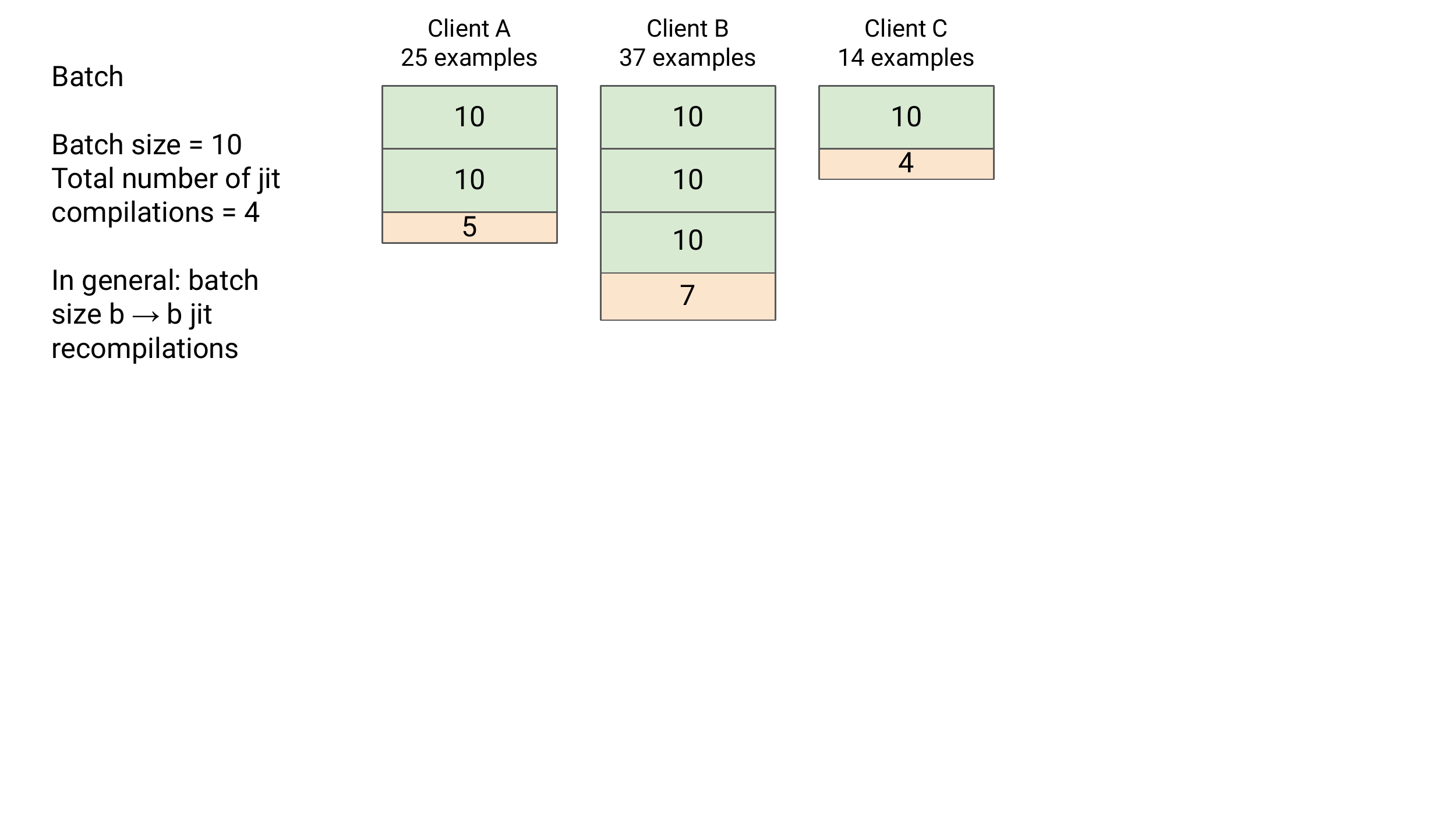}     &  \includegraphics[scale=0.35]{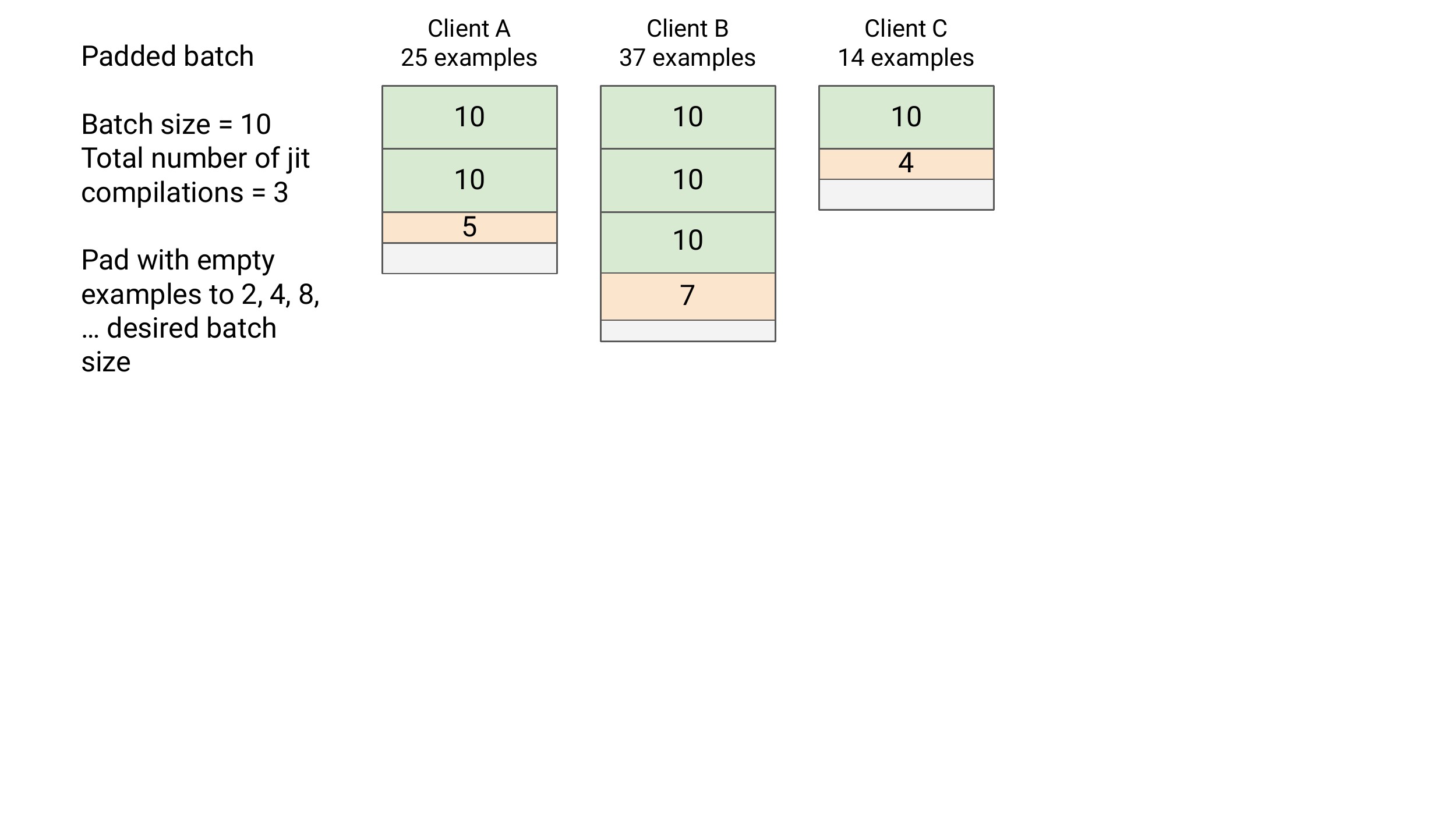} &  \includegraphics[scale=0.35]{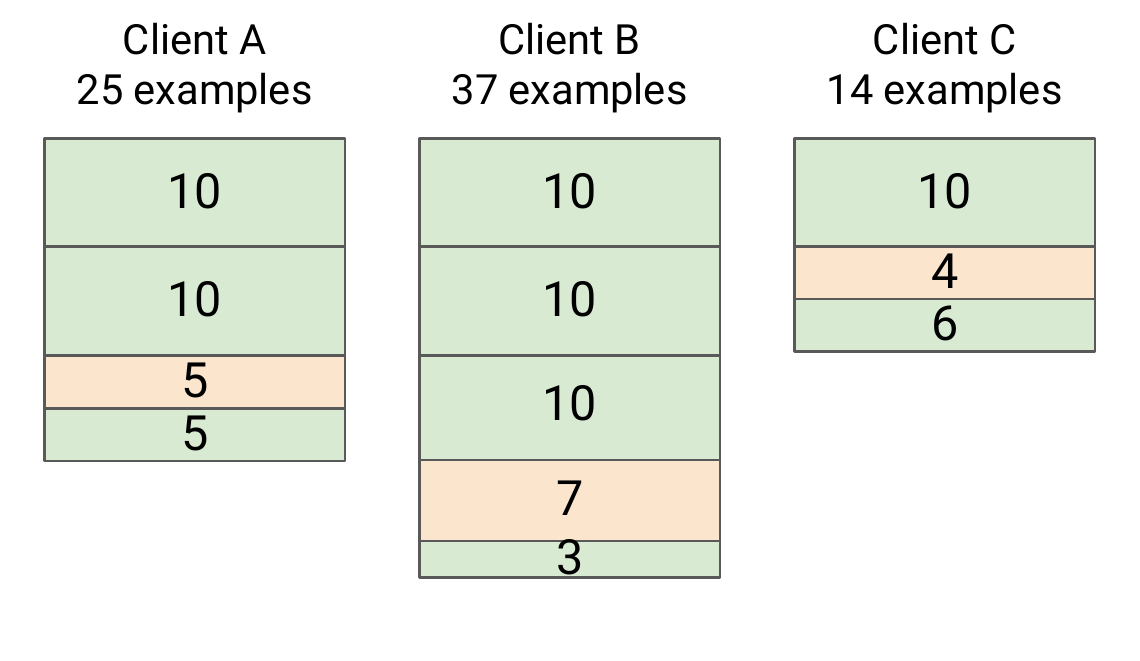} \\
    (a)    \texttt{batch} &  (b) \texttt{padded\_batch} &  (c) \texttt{shuffle\_repeat\_batch}
    \end{tabular}
    \caption{Resulting batches for three clients of various sizes with different batching strategies applied.}
    \label{fig:batching}
\end{figure}

\paragraph{JIT efficient batching}

Despite the small size of client datasets, there is no guarantee that their sizes will be the same. As a result, when we run batch evaluation on client datasets, the sizes of the final batch can vary greatly depending on the client, as shown in Figure~\ref{fig:emnist-hist}. This poses a practical challenge when using JAX. For best performance, \texttt{jax.jit}\footnote{\url{https://jax.readthedocs.io/en/latest/jax-101/02-jitting.html}} is used to perform Just In Time (JIT) compilation of a JAX Python function into XLA compiled machine code. \texttt{jax.jit} invokes the XLA compiler for each unique combination of input shapes. Left alone, the large number of possible final batch sizes results in excessive JIT recompilations, significantly slowing execution time. In response to this, we implemented three different batching strategies:
\begin{compactitem}
  \item \texttt{batch} produces batches in a fixed sequential order without padding the final batch. This is included mostly for illustration purposes.
  \item \texttt{padded\_batch} produces padded batches in a fixed sequential order for evaluation. By padding the final batch to a small number of fixed sizes, we can set a limit on the maximum number of possible JIT recompilations. This is typically used in evaluation.
  \item \texttt{shuffle\_repeat\_batch} produces batches in a shuffled and repeated order for training where shuffling is done without replacement and batches are always the same size. This is typically used in training.
\end{compactitem}
Figure~\ref{fig:batching} showcases the resulting batches using each of these batching strategies for different client dataset sizes.

\paragraph{Model and optimizer}

The model and optimizer described in this section are unchanged between the standard centralized learning setting and the federated learning setting. There are already numerous JAX based libraries for neural networks and optimizers, such as Flax, Haiku, Objax, and Optax. Thus, for convenience, \fedjax\ provides an implementation-agnostic wrapper to make porting existing models and optimizers into \fedjax\ as simple as possible. For example, using an existing Haiku model in \fedjax\ is as easy as wrapping the module in a \texttt{fedjax.Model}. \neurips{We refer readers to the tutorial\footnote{\url{https://fedjax.readthedocs.io/en/latest/notebooks/model_tutorial.html}} for an overview of these functionalities.}

\paragraph{Included datasets and models}

Currently in federated learning research, there are a variety of commonly used datasets and models, such as image recognition, language modeling, and more. A growing set of these datasets and models can be used straight out of the box in \fedjax. This not only encourages valid comparisons between various federated algorithms but also accelerates the development of new algorithms since the preprocessed datasets and models are readily available for use and do not have to be written from scratch.

At present, \fedjax\ comes packaged with the following datasets and sample models:

\begin{itemize}
    \item EMNIST-62 \citep{caldas2018leaf}, a character recognition task
    \item Shakespeare \citep{mcmahan2017communication}, a next character prediction task
    \item Stack Overflow \citep{reddi2020adaptive}, a next word prediction task
\end{itemize}

\fedjax\ also provides tools to create new datasets and models that can be used with the rest of the library along with implementations of {federated averaging} \citep{mcmahan2017communication} and other algorithms, such as {adaptive federated optimizers} \citep{reddi2020adaptive}, {agnostic federated averaging} \citep{ro2021communication}, and Mime \citep{karimireddy2020mime}.

\paragraph{Metrics}

When evaluating accuracy on a dataset, we wish to know the proportion of examples that are correctly predicted by the model. In federated learning specifically, we typically want to know the accuracy for each client and over all clients. Thus, we need to divide the work across clients, evaluate each separately, and finally somehow combine the results.


Assuming the metric evaluation logic is JIT compiled, this faces the same issue of excessive recompilations due to different subset sizes resulting in shape differences. While \texttt{padded\_batch} can be used to address this issue, it will also result in batches padded with empty examples which should not be counted in the metric calculation. In an effort to improve ease-of-use while maintaining performance, we designed \texttt{fedjax.Metric} to be defined on a single example rather than a batch of examples. This way, the metric calculation can be freely vectorized with \texttt{jax.vmap}\footnote{\url{https://jax.readthedocs.io/en/latest/jax.html?highlight=vmap##jax.vmap}} and empty examples in padded batches ignored without users having to explicitly account for this themselves.

\subsection{Client training and aggregators}

\paragraph{Client training}

As mentioned previously, federated learning experiments typically include a step of training across decentralized and distributed clients. \fedjax\ provides core functions for expressing this step. Because simulation is the focus, there is no need to introduce any design or technical overhead for distributed machine communication. Instead, the entire per-client work can run on a single machine, making the API simpler and the execution faster in most cases.

In its simplest form, a basic for-loop can be used for conducting training across multiple clients in \fedjax{}. For even faster training speeds, \fedjax\ provides the $\texttt{fedjax.for\_each\_client}$ primitive backed by $\texttt{jax.jit}$ and $\texttt{jax.pmap}$\footnote{\url{https://jax.readthedocs.io/en/latest/jax.html?highlight=pmap##jax.pmap}}, which enables the simulation work to easily run on one or more accelerators such as GPU and TPU. \neurips{We refer readers to the tutorial\footnote{\url{https://fedjax.readthedocs.io/en/latest/notebooks/algorithms_tutorial.html}} for an overview of these functionalities.} 
By defining client work in terms of \texttt{fedjax.for\_each\_client}, we are able to arbitrarily group cohorts of clients to be executed in parallel for greater performance without additional burden on the user.

\paragraph{Server aggregation}

The final step of server aggregation is often what differs the most significantly between federated learning algorithms. Thus, making this step as easily expressible and interpretable as possible is a core design goal of \fedjax. This is achieved by providing basic underlying functions for working with model parameters, which are usually structured as pytrees\footnote{\url{https://jax.readthedocs.io/en/latest/pytrees.html}} in JAX, as well as high-level pre-defined common aggregators in $\texttt{fedjax.aggregators}$. These aggregators can also be used to implement compression and differentially private federated learning algorithms.

\section{Example}
\label{sec:examples}




\begin{algorithm}[t]
\caption{Federated averaging algorithm \citep{mcmahan2017communication} }
\label{alg:FedAvg}
\begin{varwidth}[t]{0.5\textwidth}
\begin{algorithmic}[t]
\Procedure{FederatedAveraging}{}
    \State $T$: total number of rounds, $c$: number of clients per round, $\eta_s$: server learning rate.
    \State Initialize parameters: $w_0$
    \For{round $t = 1$ to $T$}
        \State $C_t \gets$ (random set of $c$ clients)
        \For{client $k \in C_t$}
        \[
        \Delta_k, n_k \gets \textsc{ClientUpdate}(k, w_{t-1})
        \]
        \EndFor
        \State $w_t \gets w_{t-1} - \eta_{\text{s}} \frac{\sum_{k \in C_t}n_k \Delta_k}{
        \sum_{k' \in C_t} n_{k'}}$
    \EndFor
\EndProcedure
\end{algorithmic}
\vspace{3ex}
\end{varwidth}
\begin{varwidth}[t]{0.5\textwidth}
\begin{algorithmic}[t]
\Procedure{ClientUpdate}{$k, w$} 
\State $S_k$: dataset of client $k$, $B$: batch size, $E$: Number of epochs, $\eta_c$: client learning rate.
\State $w' \gets w $
    \State $\mathcal{B} \gets$ (split $S_k$ into batches of size $B$)
    \For{epoch $e = 1$ to $E$}
        \For{batch $b\in \mathcal{B}$}
            $$ w' \gets w' - \eta_{\text{c}}\nabla \sum_{(x_i, y_i) \in b} L(w', x_i, y_i)$$
        \EndFor
    \EndFor
    \State \textbf{return} $w - w', |S_k|$
\EndProcedure
\end{algorithmic}
\end{varwidth}
\end{algorithm}

In this section, we demonstrate how to implement federated averaging with \fedjax. Because \fedjax\ only introduces a few core concepts and is clear and straightforward, code written in \fedjax\ tends to resemble the pseudo-code used to describe novel algorithms in research papers, making it easy to get started with. While \fedjax\ provides primitives for federated learning, they are not required and can be replaced with just NumPy and JAX. The advantage of building on top of JAX is that even the most basic implementations can still be reasonably fast.

Below, we walk through a simple example of federated averaging (Algorithm~\ref{alg:FedAvg}) for linear regression implemented in \fedjax.
The first steps are to set up the experiment by loading the federated dataset, initializing the model parameters, and defining the loss and gradient functions.
The code for these steps is given in Figure~\ref{fig:one}.

\begin{figure}[h]
\begin{tcolorbox}
\verbatimfont{\small}
\begin{verbatim}
import jax
import jax.numpy as jnp
import fedjax

# {"client_id": client_dataset}
federated_data = fedjax.FederatedData()
# Initialize model parameters
server_params = jnp.array(0.5)
# Mean squared error
mse_loss = lambda params, batch: jnp.mean(
        (jnp.dot(batch["x"], params) - batch["y"])**2)
# jax.jit for XLA and jax.grad for autograd
grad_fn = jax.jit(jax.grad(mse_loss))
\end{verbatim}
\end{tcolorbox}
\label{fig:one}
\caption{Dataset and model initialization.}
\end{figure}

Next, we use $\texttt{fedjax.for\_each\_client}$ to coordinate the training that occurs across multiple clients. For federated averaging, $\texttt{client\_init}$ initializes the client model using the server model, $\texttt{client\_step}$ completes one step of local mini-batch SGD, and $\texttt{client\_final}$ returns the difference between the initial server model and the trained client model. By using $\texttt{fedjax.for\_each\_client}$, this work will run on any available accelerators and possibly in parallel because it is backed by $\texttt{jax.jit}$ and $\texttt{jax.pmap}$. While this is already simple to implement, the same could also be written out as a basic for-loop over clients if desired. The code for these steps is given in Figure~\ref{fig:two}. This code implements the \textsc{ClientUpdate} procedure of  federated averaging from Algorithm~\ref{alg:FedAvg}.

\begin{figure}[!h]
\begin{tcolorbox}
\verbatimfont{\small}
\begin{verbatim}
# For-loop over clients with client learning rate 0.1
for_each_client = fedjax.for_each_client(
  client_init=lambda server_params, _: server_params,
  client_step=(
    lambda params, batch: params - grad_fn(params, batch) * 0.1),
  client_final=lambda server_params, params: server_params - params)
\end{verbatim}
\end{tcolorbox}
\caption{Client update.}
\label{fig:two}
\end{figure}

Finally, we run federated averaging for $100$ training rounds by sampling clients from the federated dataset, training across these clients using $\texttt{fedjax.for\_each\_client}$, and aggregating the client updates using weighted averaging to update the server model in Figure~\ref{fig:three}.  This code implements the \textsc{FederatedAveraging} procedure of  federated averaging from Algorithm~\ref{alg:FedAvg}. 

\begin{figure}[h]
\begin{tcolorbox}
\verbatimfont{\small}
\begin{verbatim}
# 100 rounds of federated training
for _ in range(100):
  clients = federated_data.clients()
  client_updates = []
  client_weights = []
  for client_id, client_update in for_each_client(server_params, clients):
    client_updates.append(client_update)
    client_weights.append(federated_data.client_size(client_id))
  # Weighted average of client updates
  server_update = (
    jnp.sum(client_updates * client_weights) /
    jnp.sum(client_weights))
  # Server learning rate of 0.01
  server_params = server_params - server_update * 0.01
\end{verbatim}
\end{tcolorbox}
\caption{Server update and the federated learning algorithm.}
\label{fig:three}
\end{figure}

\section{Benchmarks}
\label{sec:experiments}

We benchmark the \fedjax\ federated averaging implementation on the image recognition task for the federated EMNIST-62 dataset \citep{caldas2018leaf} and the next word prediction task for Stack Overflow \citep{reddi2020adaptive}.

The EMNIST-62 dataset consists of $3400$ writers and their writing samples, which are one of $62$ classes (alphanumeric).
Following \cite{reddi2020adaptive}, we train a convolutional neural network for $1500$ rounds with $10$ clients per round using federated averaging. We run experiments on GPU (a single NVIDIA V100) and TPU (a single TensorCore on a Google TPU v2) and report the final test accuracy, overall execution time, average training round duration, and full evaluation time in Table~\ref{tab:emnist-benchmarks}. We note that with a singleTensorCore, training takes under five minutes.

\begin{figure}[!h]
\centering
\includegraphics[scale=0.5]{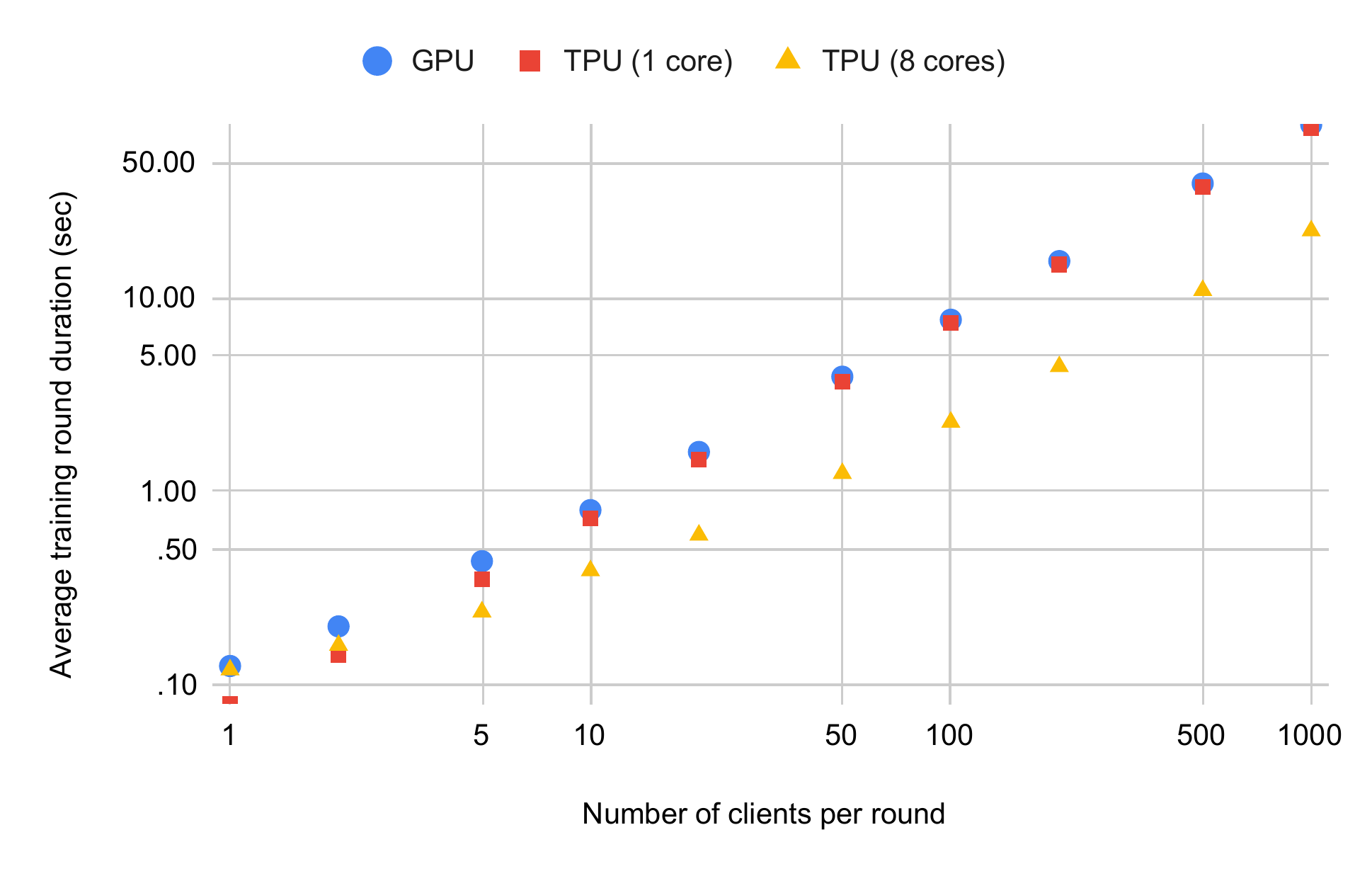}
\caption{Stack Overflow average training round duration as the number of clients per round increases.}
\label{fig:stackoverflow-clients}
\end{figure}

The Stack Overflow dataset consists of questions and answers from the Stack Overflow forum, grouped by username. This dataset consists of roughly $342$K users in the train split and $204$K in the test split. Following \cite{reddi2020adaptive}, we train a single layer LSTM for $1500$ rounds with $50$ clients per round using federated averaging. We run experiments on GPU (a single NVIDIA V100), TPU (a single TensorCore on a Google TPU v2) using only $\texttt{jax.jit}$, and multi-core TPU (eight TensorCores on a Google TPU v2) using $\texttt{jax.pmap}$ and report results in Table~\ref{tab:stackoverflow-benchmarks}. Benchmarks show that with multiple TensorCores, recurrent language models can be trained  in under an hour. Figure~\ref{fig:stackoverflow-clients} also shows the average training round duration as the number of clients per round increases. We note that training with multiple TensorCores is substantially faster as the number of clients per round increases.

\begin{table}[t]
    \centering
    \caption{Benchmark results on EMNIST with federated averaging.}
    \begin{tabular}{ccccc}
         Hardware & Test accuracy & Overall (s) & Average training round (s) & Full evaluation (s) \\
         \hline
         GPU & 85.92\% & 418 & 0.26 & 7.21 \\
         TPU & 85.85\% & 258 & 0.16 & 4.06 \\
    \end{tabular}
    \label{tab:emnist-benchmarks}
\end{table}

\begin{table}[t]
    \centering
    \caption{Benchmark results on Stack Overflow with federated averaging.}
    \begin{tabular}{ccccc}
         Hardware & Test accuracy & Overall (m) & Average training round (s) & Full evaluation (m) \\
         \hline
         GPU & 24.74\% & 127.2 & 4.33 & 17.32 \\
         TPU ($\texttt{jax.jit}$) & 24.44\% & 106.8 & 3.73 & 11.97 \\
         TPU ($\texttt{jax.pmap}$) & 24.67\% & 48.0 & 1.26 & 11.97 \\
    \end{tabular}
    \label{tab:stackoverflow-benchmarks}
\end{table}

\section{Conclusion}

We introduced \fedjax, a JAX-based open source library for federated learning simulations that emphasizes ease-of-use in research.
\fedjax\ provides simple primitives for federated learning along with a collection of canonical datasets, models, and algorithms to make developing and evaluating federated algorithms easier and faster. Implementing additional algorithms, datasets, and models remains an on-going effort.

\neurips{\begin{ack}
Authors thank Sai Praneeth Kamireddy for contributing to the library and various discussions early on in development.

Authors also thank Ehsan Amid, Theresa Breiner, Mingqing Chen, Fabio Costa, Roy Frostig, Zachary Garrett, Satyen Kale, Rajiv Mathews, Lara Mcconnaughey, Brendan McMahan, Mehryar Mohri, Krzysztof Ostrowski, Max Rabinovich, Michael Riley, Gary Sivek, Jane Shapiro, Luciana Toledo-Lopez, and Michael Wunder for helpful comments and contributions.
\end{ack}}

\arxiv{
\section*{Acknowledgements}
Authors thank Sai Praneeth Kamireddy for contributing to the library and various discussions early on in development.

Authors also thank Ehsan Amid, Theresa Breiner, Mingqing Chen, Fabio Costa, Roy Frostig, Zachary Garrett, Satyen Kale, Rajiv Mathews, Lara Mcconnaughey, Brendan McMahan, Mehryar Mohri, Krzysztof Ostrowski, Max Rabinovich, Michael Riley, Gary Sivek, Jane Shapiro, Luciana Toledo-Lopez, and Michael Wunder for helpful comments and contributions.
}

\bibliographystyle{plainnat}
\bibliography{references}

\end{document}